%% file: main.tex
\documentclass{article}

\usepackage{microtype}
\usepackage{graphicx}
\usepackage{subfigure}
\usepackage{booktabs} 

\usepackage{hyperref}

\usepackage[accepted]{icml_r2fm}

\usepackage{amsmath}
\usepackage{amssymb}
\usepackage{mathtools}
\usepackage{amsthm}
\usepackage{enumitem}
\usepackage{caption}
\usepackage{subcaption}

\usepackage[capitalize,noabbrev]{cleveref}

\theoremstyle{plain}

\theoremstyle{definition}

\theoremstyle{remark}

\usepackage[textsize=tiny]{todonotes}

\icmltitlerunning{Do Sparse Autoencoders Generalize?}

\begin{document}

\twocolumn[
    \icmltitle{Do Sparse Autoencoders Generalize? A Case Study of Answerability}
    
    
    
    \icmlsetsymbol{equal}{*}
    \icmlsetsymbol{equaladv}{*}

    \begin{icmlauthorlist}
    \icmlauthor{Lovis Heindrich}{work}
    \icmlauthor{Philip Torr}{oxf,box}
    \icmlauthor{Fazl Barez}{oxf,equaladv}
    \icmlauthor{Veronika Thost}{ibm,equaladv}
    \end{icmlauthorlist}

    \icmlaffiliation{work}{Work done during a research visit at University of Oxford.}
    \icmlaffiliation{oxf}{University of Oxford, Oxford, United Kingdom}
    \icmlaffiliation{box}{WhiteBox}
    \icmlaffiliation{ibm}{MIT-IBM Watson AI Lab}
    
    \icmlcorrespondingauthor{Lovis Heindrich}{lovis.heindrich@student.uni-tuebingen.de}
    
    \icmlkeywords{mechanistic interpretability, sparse autoencoders, robustness, out-of-distribution evaluation, linear probes, answerability}
    
    \vskip 0.3in
]



\printAffiliationsAndNotice{\icmlEqualContribution} 

\begin{abstract}
Sparse autoencoders (SAEs) have emerged as a promising approach in language model interpretability, offering unsupervised extraction of sparse features. For interpretability methods to succeed, they must identify abstract features across domains, and these features can often manifest differently in each context. We examine this through "answerability"—a model's ability to recognize answerable questions. We extensively evaluate SAE feature generalization across diverse, partly self-constructed answerability datasets for Gemma 2 SAEs. Our analysis reveals that residual stream probes outperform SAE features within domains, but generalization performance differs sharply. SAE features show inconsistent out-of-domain transfer, with performance varying from almost random to outperforming residual stream probes. Overall, this demonstrates the need for robust evaluation methods and quantitative approaches to predict feature generalization in SAE-based interpretability.

\end{abstract}

\input{latex/01-intro}
\input{latex/03-related}
\input{latex/04-method}

\input{latex/05-evaluation}
\input{latex/0-6conclusions}
\input{latex/limitations}
\input{latex/ethics}



\bibliography{main}
\bibliographystyle{icml2025}

\appendix
\label{sec:appendix}
\input{latex/07-appendix}

\end{document}

%% file: latex/01-intro.tex
\section{Introduction}

Language models increasingly drive real-world applications, yet their black-box nature remains a fundamental barrier to deployment. This lack of visibility has sparked intense interest in interpretability methods \cite{olah2018building}, with sparse autoencoders (SAEs) emerging as a particularly promising direction \cite{cunningham2023sparse, bricken2023monosemanticity}. The core idea is appealingly simple: train an autoencoder to reconstruct neural activations through a sparse bottleneck, forcing the model to learn disentangled, interpretable features.

\begin{figure}
    \centering
    \includegraphics[width=0.99\linewidth,trim={0 2.3cm 0 3.4cm},clip]{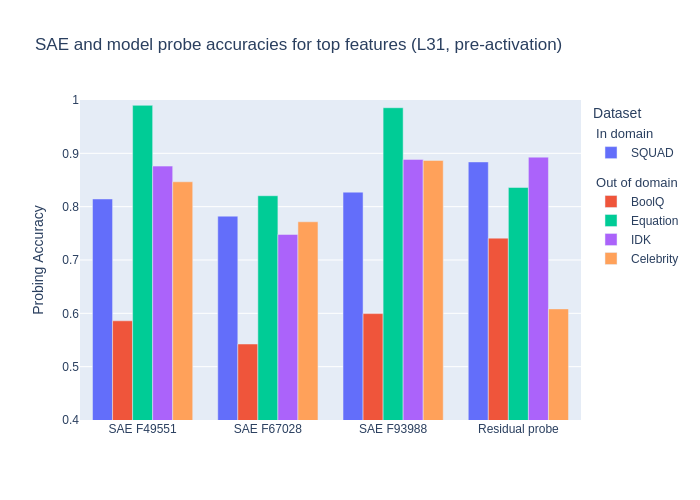}
    \caption{SAE features vs linear probes trained on SQUAD answerability data: generalization performance varies drastically with different answerability OOD datasets. While the features generalize well to some datasets (Equation, IDK, Celebrity), generalization to BoolQ is almost random.}
    \label{fig:enter-label}
\end{figure}

Recent work demonstrates that SAEs can effectively capture 
a wide range of features in language models such as errors in code, sycophancy, and gender bias \cite{templeton2024scaling}, as well as syntax patterns and sentiment \cite{lan2024sparse, marks2024enhancing}. In parallel, circuit-level analyses have begun to reveal mechanistic underpinnings of neural network behavior \citep{elhage2021mathematical, olsson2022context, marks2024sparse}.
 While interpretability methods like SAEs are often motivated by AI safety concerns, recent work suggests that even advanced interpretability approaches may have fundamental limitations for ensuring AI safety \cite{Barez2025-open}.
Despite these advances and critiques, a crucial question remains unaddressed: can these methods fully capture the abstract concepts that would help us understand how language models think and process information in a robust way?
We study this question by focusing on answerability—a model's ability to recognize whether it can answer a question. This capability is fundamental to language model behavior and exists across diverse tasks. If SAEs truly capture meaningful abstractions, they should be able to extract features representing this capability.
Specifically, we evaluate how well a model's notion of answerability generalizes across different datasets by (1) training answerability probes on an in-domain dataset and (2) evaluating the probes on four out-of-distribution datasets, including two synthetic datasets designed by us.
Our experimental results reveal a more nuanced picture than previous work suggests. While residual stream probes outperform SAE features within specific domains, we obtain more mixed results
when we look at generalization. \textbf{We find good SAE features for individual domains, but varying transfer abilities across datasets. Similarly, residual stream probes exhibit high variance in generalization despite strong in-domain performance}. These findings raise important questions about SAE research, particularly for abstract or more complex concepts that manifest differently across contexts.

%% file: latex/03-related.tex
\section{Preliminaries}

Sparse autoencoders (SAE) have recently been proposed as an interpretability method that aims to decompose the dense and usually uninterpretable activations $\mathbf{x}$ of an LLM into a sparse and interpretable representation $\mathbf{f}$ \cite{cunningham2023sparse, bricken2023monosemanticity,yun2021transformer}. This is achieved by training an autoencoder to reconstruct model activations, such that the learned representation is significantly wider and sparser than the original activations. 

A standard formulation of the autoencoder consists of an encoder and decoder layer with a ReLU activation function (Equations~\eqref{eq:sae1} and~\eqref{eq:sae2}, \citealp{bricken2023monosemanticity}), where $W_e\in\mathbb{R}^{d_{sae}\times d_{model}}$ with $d_{sae}\gg d_{model}$ is the encoder layer with bias term $\mathbf{b}_e$, and $W_d$ is the decoder layer with bias term $\mathbf{b}_d$. 

\begin{align}
    \mathbf{f}&=\mathrm{ReLU}(W_e (\mathbf{x}-\mathbf{b}_d) +\mathbf{b}_e) \label{eq:sae1} \\
    \hat{\mathbf{x}} &= W_d\mathbf{f} + \mathbf{b}_d \label{eq:sae2} 
\end{align}

While initially sparsity has been achieved through L1 regularization in the loss function (Equation~\eqref{eq:sae_loss}, \citealp{bricken2023monosemanticity}), recent research on SAEs has focused on designing improved methods to achieve sparsity \cite{lieberum2024gemma, gao2024scaling}.  
\begin{align}
    \mathcal{L} &= \left\| \mathbf{x} - \hat{\mathbf{x}} \right\|_2^2 + \lambda \left\| \mathbf{f} \right\|_1 \label{eq:sae_loss}
\end{align}

\section{Related Works}

\paragraph{SAE Training}
Much SAE research is dedicated to improving training efficiency and effectiveness, but the latter is usually measured in terms of reconstruction quality and hence disconnected from downstream scenarios \cite{rajamanoharan2024improving,lieberum2024gemma}. 
Only most recent work addresses this issue.
\citet{gao2024scaling} apply downstream probing, yet the classification tasks considered are most simple (e.g sentiment, language identification) and likely do not test any generalization. 
Similarly, \citet{makelov2024towards} use downstream data without considering generalization.


\paragraph{SAE Downstream Evaluation}
Several studies evaluate SAE features, including downstream settings.
Yet, research often focuses on simple syntactic features or does not evaluate how general the discovered features are
\cite{yun2021transformer,bricken2023monosemanticity,kissane2024interpreting}.
%
There are various works focusing on more abstract concepts such as 
indirect object identification \cite{cunningham2023sparse} and subject-verb agreement \cite{marks2024sparse},
but those are still directly related to syntax. In contrast, answerability often depends on domain-specific background knowledge (e.g., math or factual knowledge) and hence better suits the study on generalization.
%
\citet{demircan2024sparse} consider the representation of quality estimates from reinforcement learning, and hence a rather complex concept, but they focus on task-specific (vs generalizable) SAEs.

Closest to our work are the following studies.
%
\citet{bricken2024using} focus on comparing SAE features to linear probes as bioweapon classifiers.
Similarly to us, they show that the SAE probes are competitive but more complex and brittle; for instance, already format mismatch between the transformer/SAE/probe training data may degrade performance. When evaluated on multilingual out-of-distribution data (similar to in-domain data but in different languages), they find that SAE features can generalize well in specific settings in the mostly lexical task of bioweapon classification. 
%
\citet{kantamneni2024saeprobing,kantamneni2025sparseautoencodersusefulcase} similarly conduct experiments comparing to traditional probing on activations and demonstrate that SAEs seem to work in certain scenarios (e.g., with very small datasets or corrupted data). They also consider out-of-distribution data, yet focus more on multilingual and syntactic variations. 
Similar to us, they obtain mixed results without clear conclusion which probes are better. Generally SAEs do not outperform regular probing.
Our evaluation lifts this work to 
a more complex task and a greater variety of distributions.

In summary, 
prior work has largely neglected generalization beyond multilingual/syntactic scenarios. Our study closes this gap in the context of a suitable, complex concept, answerability, which likely manifests differently across contexts.

%% file: latex/04-method.tex
\begin{figure}[ht!]
    \centering
\includegraphics[width=0.99\linewidth,trim={0 0 0 0},clip]{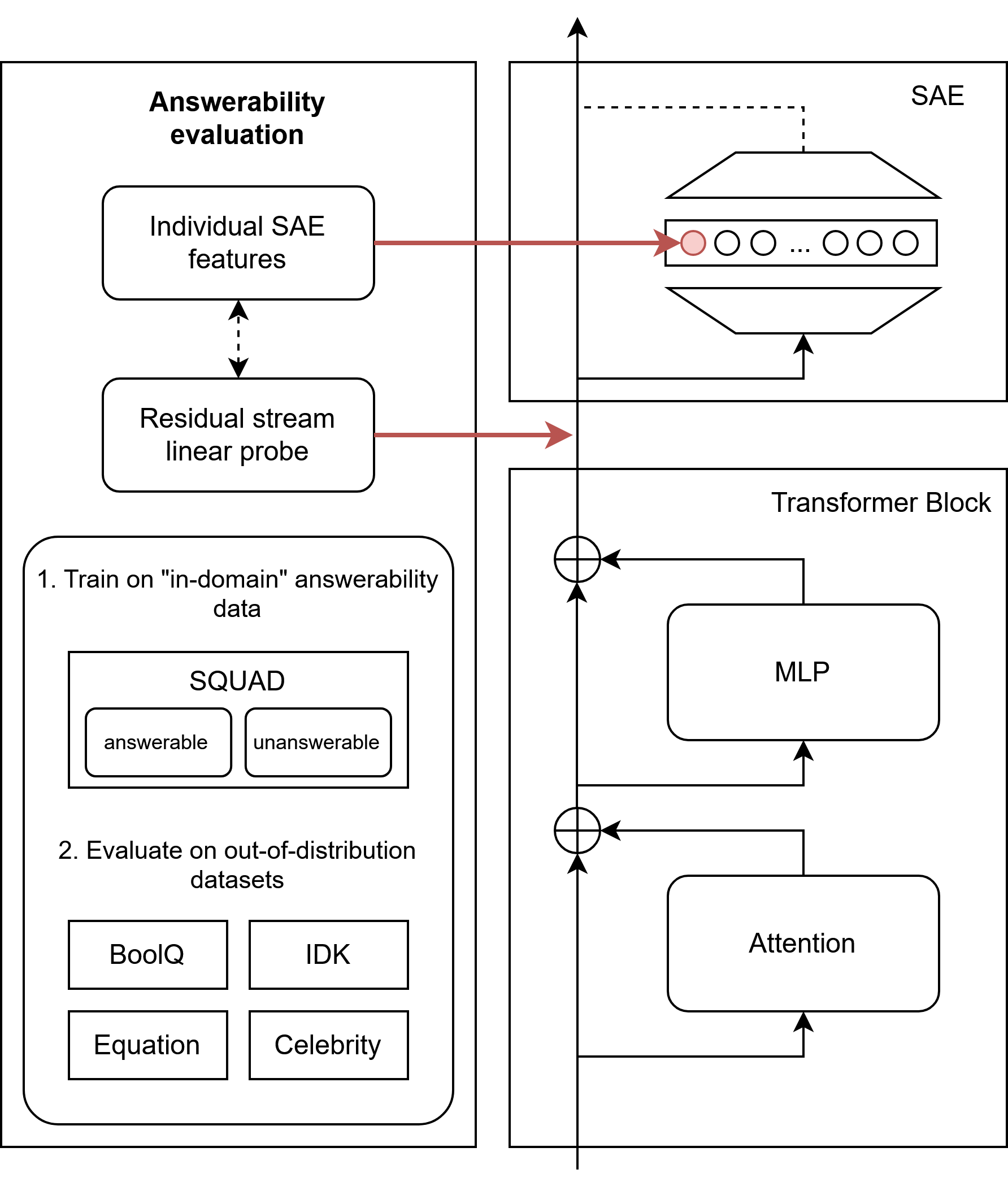}
    \caption{Visualization of our evaluation approach on a simplified transformer and SAE architecture. Activations from individual SAE features and the full residual stream are used to train answerability probes on SQUAD. Generalization performance is evaluated on multiple OOD datasets.}
    \label{fig:method-summary}
\end{figure}

\section{Methodology}

We evaluate SAE probes for answerability detection with a specific focus on  generalization. To evaluate generalization of individual SAE features, we select features which activations lead to high answerability detection on the training dataset, and evaluate if the same features can also detect answerability on a range of out-of-distribution datasets. We compare SAE feature generalization to a baseline of training a linear probe on the residual stream of the LLM. Figure~\ref{fig:method-summary} provides an overview of the methodology.

\paragraph{SAE Probes}
We use the "Gemma Scope" SAEs pretrained by \citet{lieberum2024gemma} for the instruction-tuned model Gemma 2 \citep{team2024gemma}, and specifically the largest available SAEs with a width (number of dimensions) of 131k. \citet{lieberum2024gemma} provide SAEs trained on layers 20 and 31.
Note that answerability more generally (i.e., beyond specific types of answerability) is a rather high-level concept, which we assume to be represented in intermediate and later layers.

We search for SAE features using 2k samples of SQUAD (balanced, leaving 1.8k for testing). We collect the feature activations on the last token position and then use 5-fold cross validation for finding SAE features that are predictive for answerability, thus obtaining 1-sparse SAE probes \citep{gurnee2023finding}.
We then train final probes\footnote{We use SAE probes and SAE features synonymously.} (i.e., scale and bias) for best performing features, which are used for the out-of-distribution evaluation. See Appendix~\ref{app:prelims} for details. ``Top'' features are selected based on training set performance.

\paragraph{Baselines: Linear Probes}
We train simple linear residual stream probes on the (in-domain) training dataset we also use for finding the SAE features. To ensure robustness, we employ bootstrap analysis across different training splits. 
Since we also focus on SAE features for the residual stream, this probing represents an upper bound for the SAE probing performance on in-domain data.
Observe that these probes achieve 85-90\% accuracy on the in-domain SQUAD data, and thus provide a strong benchmark for comparison. 

We do not evaluate the SAE or residual stream probes against alternative methods for answerability detection (e.g. \citealp[]{chuang2024lookback}), as our goal was to evaluate SAE generalization capabilities, instead of developing a novel method for answerability detection. 

\begin{table}[hb!]
    \centering
        \caption{Number of total examples per dataset. The number of answerable and unsanwerable examples is balanced within each used dataset.}
    \begin{tabular}{lc}
        \hline
        Dataset & Size \\
        \hline
        SQUAD (train) & 2000 \\
        BoolQ (train) & 2000 \\
        SQUAD (test) & 1800 \\
        SQUAD (variations) & 1800 \\
        BoolQ (test) & 2000 \\
        IDK & 484 \\
        Equation & 2000 \\
        Celebrity & 600 \\
        \hline
    \end{tabular}
    \label{table:dataset-size}
\end{table}%

\begin{table*}[ht!]
\centering
\caption{Answerable and unanswerable examples from the used datasets.}
\small 
\setlength{\tabcolsep}{4pt} 
\begin{tabular}{l p{5.5cm} p{5.5cm}}
    \toprule
    \textbf{Dataset} & \textbf{Answerable} & \textbf{Unanswerable} \\
    \midrule
    SQUAD & 
    \textbf{Passage:} The first beer pump known in England is believed to [\dots]. 
    \newline 
    \textbf{Question:} When was John Lofting born? 
    & 
    \textbf{Passage:} Starting in 2010/2011, Hauptschulen were merged  [\dots]. 
    \newline 
    \textbf{Question:} In what school year were Hauptschulen last combined with Realschulen and Gesamtschulen? \\
    \midrule
    IDK & 
    \textbf{Passage:} Singapore has reported 16 deaths. \newline 
    \textbf{Question:} Where are the deaths? 
    & 
    \textbf{Passage:} Showed the arrest of the prime suspect. \newline 
    \textbf{Question:} Where was the arrest? \\
    \midrule
    BoolQ & 
    \textbf{Passage:} On April 20, 2018, ABC officially renewed \textit{Grey's Anatomy} for a network primetime drama record-tying fifteenth season. \newline 
    \textbf{Question:} Is season 14 the last of \textit{Grey's Anatomy}? 
    & 
    \textbf{Passage:} Discover is the fourth largest credit card brand in the U.S., behind Visa, MasterCard, and American Express, with nearly 44 million cardholders. \newline 
    \textbf{Question:} Are pasilla chiles and poblano chiles the same? \\
    \midrule
    Equation & 
    \textbf{Given equations:} \newline n = 53 \newline v = 90 \newline 
    \textbf{Final equation:} \newline n / v = 
    & 
    \textbf{Given equations:} \newline n = 17 \newline u = 38 \newline 
    \textbf{Final equation:} \newline n * t = \\
    \midrule
    Celebrity & 
    \textbf{Article:} Yesterday, I saw an article about Gerard Butler. They really are a great actor. \newline 
    \textbf{Question:} Do you know what their age is? 
    & 
    \textbf{Article:} Yesterday, I saw an article about Tania Scott. They really are a great actor. \newline 
    \textbf{Question:} Do you know what their age is? \\
    \bottomrule
\end{tabular}
\label{tab:answerability}
\end{table*}
\paragraph{Datasets}
We focus on context-based question answering in the English language; we further present a short experiment using Spanish language test data in Appendix~\ref{sec:spanish_eval}. Unless otherwise specified, we treat SQUAD as the in-domain dataset used for training linear probes, and treat all other datasets as out-of-domain (OOD) test datasets. 
We use established data as well as datasets specifically constructed  for out-of-distribution evaluation; for examples see Table~\ref{tab:answerability}. Table~\ref{table:dataset-size} further shows the number of examples for each dataset used in our evaluation.

\begin{itemize}
    \item \textbf{SQUAD} \citep{rajpurkar2018know}: Dataset consisting of a short context passage and a question relating to the context. We follow the training data split and prompting template provided by \citet{slobodkin2023curious}.
    \item \textbf{IDK} \citep{sulem2021we}: Dataset with questions in the style of SQUAD, containing both answerable and unanswerable examples. We specifically use the non-competitive and unanswerable subsets of the ACE-whQA dataset.
    \item \textbf{BoolQ\_3L} 
    \citep{sulem2022yes}: Yes/no questions with answerable and unanswerable subsets.
    \item \textbf{Math Equations}: Synthetic dataset contrasting solvable equations with equations containing unknown variables.
    \item \textbf{Celebrity Recognition}: Context-based queries requiring knowledge about celebrities.
    For construction, we use a public dataset of actors and movies from IMDB\footnote{\url{https://www.kaggle.com/datasets/darinhawley/imdb-films-by-actor-for-10k-actors}}, and generate a list of the 1000 most popular actors after 1990, as measured by the total number of ratings their movies received. We construct an additional dataset of non-celebrity names by randomly generating first and last name combinations using the most common North American names from Wikipedia\footnote{\url{https://en.wikipedia.org/wiki/Lists_of_most_common_surnames_in_North_American_countries} and \url{https://en.wikipedia.org/wiki/List_of_most_popular_given_names?utm_source=chatgpt.com}}. 
\end{itemize}

%% file: latex/05-evaluation.tex
\begin{figure*}[h!]
    \centering 
     \includegraphics[width=.5\textwidth,trim={0 1.4cm 0 3.4cm},clip]{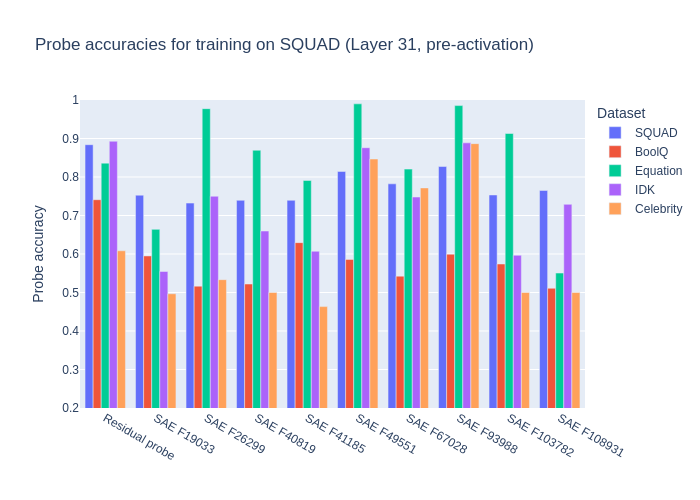}
    ~%
    \includegraphics[width=.5\textwidth,trim={0cm 1.4cm 0 3.4cm},clip]{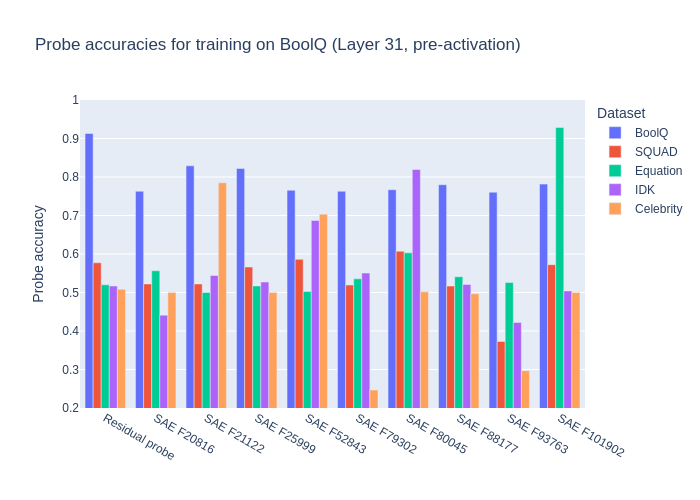}
\caption{Out-of-distribution comparison between the top SAE features and the linear probe on layer 31. SAE features are trained from pre-ReLU activations. Two in-domain training datasets are evaluated, SQUAD (left) and BoolQ (right), while all other datasets are used as OOD data.}
    \label{fig:sae-probe_pre31}
\end{figure*}

\begin{figure}[h!]
    \centering 
    \includegraphics[width=.5\textwidth,trim={0 1.4cm 0 3.4cm},clip]{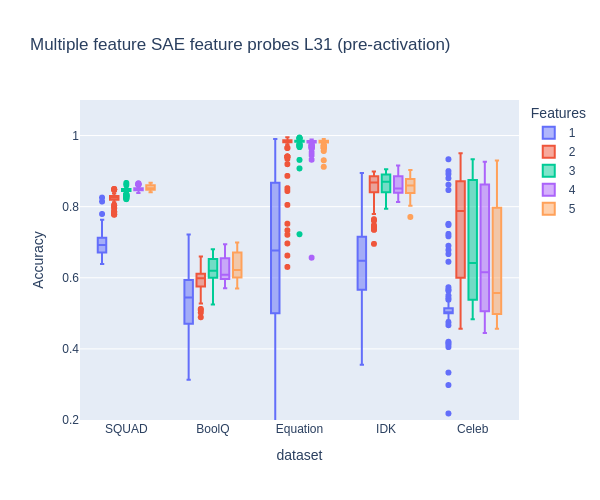}
    \caption{OOD generalization performance for combinations of multiple top SAE features, with different amounts of features (1 to 5). The median value across top feature groups with quartile ranges in the error bars is shown.}
    \label{fig:sae-k-pre31}
\end{figure}

\section{Evaluation}

In the following, we present of our main experiments; see Appendix~\ref{app:additional-analysis} for additional findings.


\paragraph{Linear vs SAE Probes: Generalization, Figure~\ref{fig:sae-probe_pre31}} 
We focus first on layer 31.
In domain, the best SAE features reach an accuracy of around 0.8 while the linear probes reach 0.9. Note that this is not surprising, since the probes have more parameters that are actually trained and thus optimized for this data. Nevertheless, it shows some advantage of probes in case in-domain data is available. 

We see rather great variation across our out-of-distribution datasets. Our custom Equation data stands out in that several SAE features and also the probe reach high performance. This seems to show that the mathematical context makes answerability easier to detect in general (as top accuracies are very high), and that the internal representation of answerability between the two datasets is compatible (some of the SAE features selected for high answerability detection on SQUAD also detect answerability on the Equation data). Notably, the best SAE features generalize better to Equation data than the residual stream probe. This only applies to the features in layer 31 and trained on SQUAD, the generalization performance is considerably worse on layer 20, see Figure~\ref{fig:sae-probe_pre20} in Appendix~\ref{app:eval-other-layers}. 

Similarly, some, but few, top SAE features reach considerable performance out of distribution on IDK (matching the performance of the linear probe) and the Celebrity data (outperforming the linear probe). Yet, the generalization performance on BoolQ is considerably bad. 
On the other hand, the linear probe performs bad on Celeb. Figure~\ref{fig:res-probe-all} in Appendix~\ref{app:eval-other-layers} shows the median value over 10 bootstrap samples including quartiles in the error bars; overall it correlates with performance.

These discrepancies in generalization can not be explained by the overall difficulty of the datasets, when selecting top SAE features for each dataset individually (i.e. training the 1-sparse probes on each dataset instead of evaluating the 1-sparse probes trained on SQUAD), we detect high performing features for each dataset (see Figure~\ref{fig:top-in-domain} in the Appendix). Instead, the results indicate that the internal representation of answerability varies between the different datasets, such that neither SAE features nor the residual stream probes represent the concept in a fully generalizable way.

Prior work \cite{sclar2023quantifying} found that LLMs can be highly sensitive to minor variations in their prompts. We evaluated generalization performance on small prompt variations of the SQUAD dataset (see Table~\ref{table:variations}) as additional out-of-distribution datasets. Figure~\ref{fig:prompt-variation} in Appendix~\ref{app:prompt-variations} shows that while both SAE features and linear probes often fail to generalize to our true OOD datasets, they generalize well to this much easier case of out-of-distribution data.

We repeated our analysis for layer 20 of Gemma 2 (Figure~\ref{fig:sae-probe_pre20}) and generally observe worse performance. 
Interestingly, the numbers for Celeb are significantly worse than all others for both the SAE and the linear probes. Since we see one exception (an SAE feature with higher than random performance), we hypothesize that there are special features encoding knowledge about celebrities which do not happen to be among our top answerability features. 
In fact, a closer investigation reveals that there are good features for BoolQ and our domain-specific Equation and Celeb datasets on layer 20 already (see Figure~\ref{fig:top-in-domain} in Appendix~\ref{app:in-domain-eval}), but they are not the same features as the ones found by training on SQuAD. 

Finally, we confirmed our findings by also training our probes on  BoolQ (also 2k samples) and evaluating generalization on the other datasets. We mainly see that varying the training data can make generalization considerably worse, even when training on the same task with seemingly similar, but potentially lower-quality data. The unanswerable samples in BoolQ were constructed by combining contexts and questions of similar dataset samples, and therefore capture only one type of unanswerability.

Overall, our experiments demonstrate one main critical issue with OOD data: \emph{the standard procedure for finding good SAE features can easily fail to select features that generalize, even if good features are available}. 
The fact that good features exist while the linear probes also fail shows some potential of SAEs. Yet finding good, generalizing features represents an open challenge.


\paragraph{Top Features, 
Figure~\ref{fig:sae-probe_pre31}} 
Interestingly, the top three features on the in-domain SQuAD data happen to also generalize better here. While the in-domain performance on SQUAD of the top SAE features is lower than the residual-stream probe, some of the generalization performance surpasses not only the probe's generalization but also the in-domain performance, indicating that some SAE features represent more general concepts that also detect answerability on the OOD data. 

These results do not hold  beyond the top-1 feature more generally, see Appendix~\ref{app:eval-other-layers}. For BoolQ, the variability of the results precludes clear conclusions.

\paragraph{SAE Feature Combinations, Figure~\ref{fig:sae-k-pre31}} 
Given the partly domain-specific nature of our out-of-distribution datasets, we hypothesized that combinations of features might work better as general probes. 
However, while increasing the number of SAE features improves the in-domain performance, OOD performance doesn't improve upon the best performing individual feature (top of blue error bars) here; layer 31, pre-activation. Other examples in Appendix~\ref{app:feature-combinations} show similar trends, and even some degradation. This underlines our above finding that the OOD setting requires better methods for SAE feature search, potentially in combination with new SAE training methods that encourage general features to be learned.


\begin{figure}[h!]
    \centering
    \begin{minipage}[b]{0.15\textwidth}
        \centering
        \includegraphics[width=\textwidth,trim={52mm 41mm 0 34mm},clip]{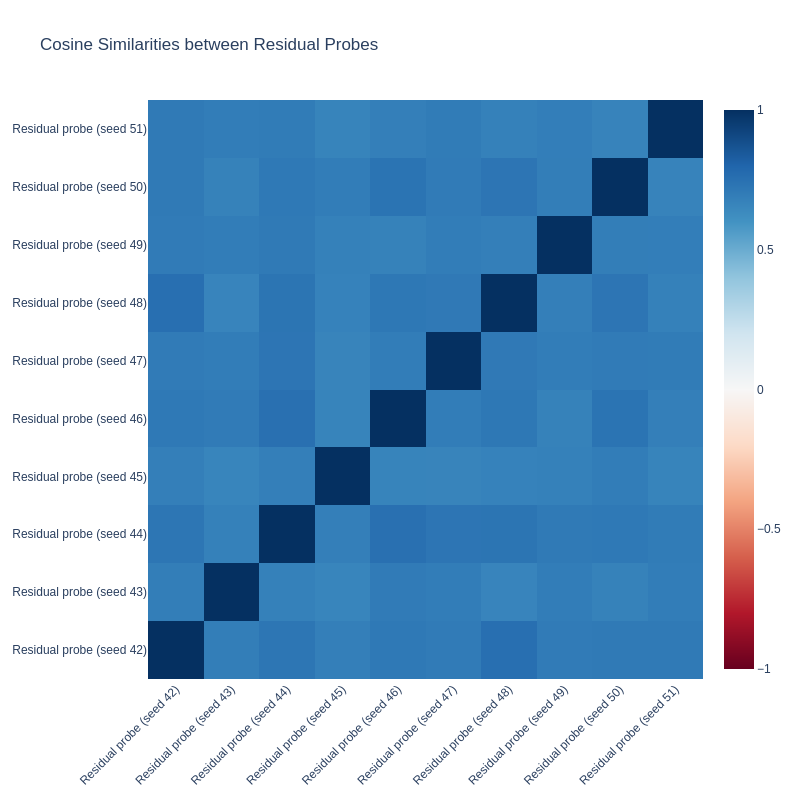}
        \caption*{(a) Residual probe similarities}
    \end{minipage}
    \hfill
    \begin{minipage}[b]{0.15\textwidth}
        \centering
        \includegraphics[width=\textwidth,trim={44mm 36mm 0 34mm},clip]{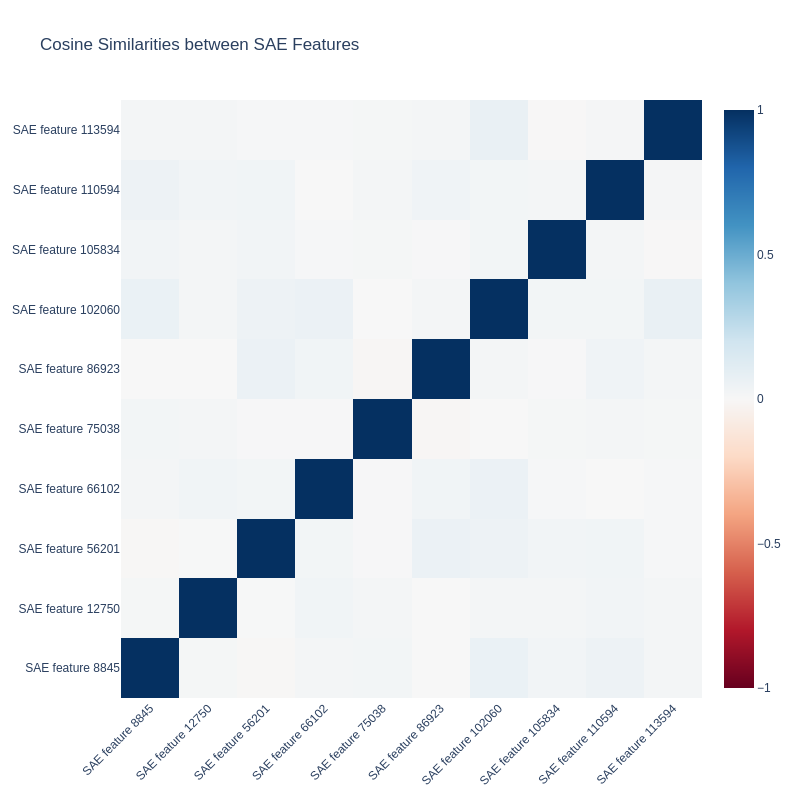}
        \caption*{(b) SAE feature similarities}
    \end{minipage}
    \hfill
    \begin{minipage}[b]{0.15\textwidth}
        \centering
        \includegraphics[width=\textwidth,trim={44mm 41mm 0 34mm},clip]{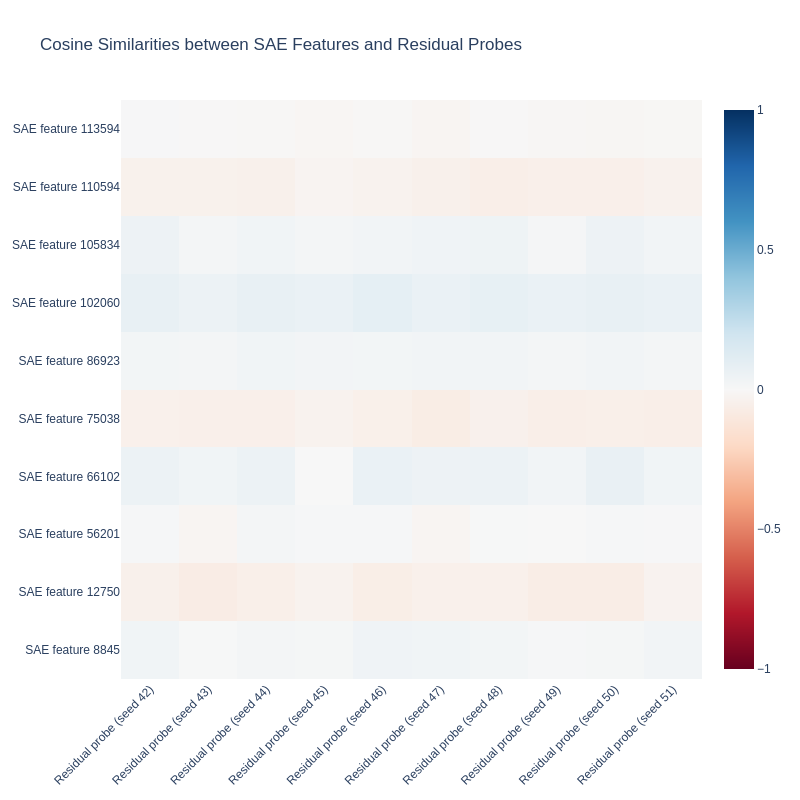}
        \caption*{(c) SAE residual similarities}
    \end{minipage}
    \caption{Pairwise cosine similarities of (a) ten residual stream probes across different seeds, (b) top SAE features and (c) top SAE features (y-axis) with linear probes across training seeds (x-axis).}
    \label{fig:similarities}
\end{figure}
  
\paragraph{Feature Similarity, Figures~\ref{fig:similarities} \& \ref{fig:similarity_k}}

We investigated whether how similar individual SAE features with high answerability performance are to each other or to the residual stream probes with cosine similarities. With an average similarity of 0.7, we find that different residual stream probes learn highly similar features, but different high performing SAE features are highly dissimilar from each other and also highly different from the linear probes. Interestingly, the best generalizing SAE feature turns out to have the highest (though overall very low) similarity with the probes. Figure~\ref{fig:similarity_k} in Appendix~\ref{app:cosine-similarities} shows that averaging multiple high performing SAE features yields greater similarity with linear probes.


%% file: latex/0-6conclusions.tex
\section{Conclusions}

We extensively evaluated SAE features for Gemma~2 in the out-of-distribution scenario using a variety of established and custom datasets. On the bright side, we find good SAE features for answerability across these domains. However, we show from various angles that the standard SAE feature search can fail to identify generalizing features, even if they exist. Additionally, fully general features might not be learned by standard SAE training methods, limiting their potential applications for steering and probing.
We hypothesize that this is due to both suboptimal training objectives and feature splitting with complex concepts \cite{bricken2023monosemanticity, chanin2024absorption}.
This shows the need for better technology for evaluating SAE features before SAEs are robustly applicable in practice.
Overall, our results indicate that comprehensive out-of-distribution evaluation is critical for future SAE-based interpretability applications. 


%% file: latex/limitations.tex
\section*{Limitations}
\label{sec:limitations}

Our evaluation is bottlenecked by the availability of high-quality SAEs trained on instruction-tuned LLMs. Therefore, we could only evaluate generalization on Gemma 2, where open-source SAEs for are available \cite{lieberum2024gemma}. While there are available SAEs for other open-source LLMs like Llama 3 \citep{dubey2024llama}, these are generally trained on the base models or of uncertain quality. In preliminary tests with Llama 3, we could not find usable answerability features for a probing comparison. 
Furthermore, the SAE features highly depend on the training hyperparameters and data they were trained on. Nevertheless, we would expect certain features to exist when aiming to use such SAEs in practice.
Finally, we used rather simple techniques for SAE probe training, which did not yield best results. But this is the point we intend to make in this work, that the existing technology is insufficient in detecting generalizable SAE features.
A key issue for SAE probing could be feature splitting, a phenomenon where SAEs of different sizes learn features in different granularities, often splitting more general features into multiple more specialized features \citep{marks2024sparse, chanin2024absorption}. If abstract concepts like answerability are split into many separate features, this can cause problems for feature-based practical applications.
\section*{Acknowledgements}
Lovis Heindrich's work on this project was partially funded through a Manifund AI Safety grant. \\
We thank TVG interns and members—particularly Minseon Kim, Luke Marks, Clement Neo, Michael Lan, and Tingchen Fu—for weekly discussions and conversations about AI safety and interpretability. Lovis thanks Joseph Bloom for useful discussions. \\
The authors also thank the developers of TransformerLens \cite{nanda2022transformerlens} and SAELens \cite{bloom2024saetrainingcodebase}, open-source libraries for mechanistic interpretability. 

%% file: latex/ethics.tex








%% file: latex/07-appendix.tex
\input{latex/02-preliminaries}

\section{Additional analysis} \label{app:additional-analysis}

\subsection{Answerability Detection at Different Layers} \label{app:eval-other-layers}
\begin{figure*}[t]
    \centering
    \includegraphics[width=0.8\textwidth,trim={0 1.4cm 0 3.4cm},clip]{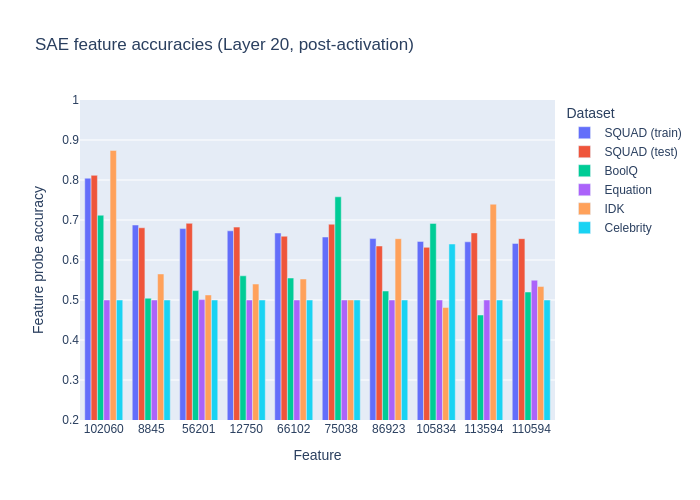}
    \caption{Answerability detection accuracies for top SAE features (Layer 20, post-activation).}
    \label{fig:sae-probe_post20}
\end{figure*}

\begin{figure*}[t]
    \centering
    \includegraphics[width=0.8\textwidth,trim={0 1.4cm 0 3.4cm},clip]{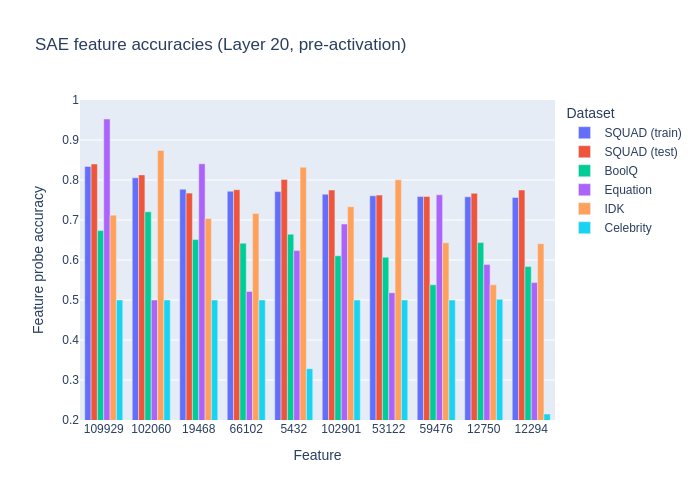}
    \caption{Answerability detection accuracies for top SAE features (Layer 20, pre-activation).}
    \label{fig:sae-probe_pre20}
\end{figure*}

\begin{figure*}[t]
    \centering
    \includegraphics[width=0.8\textwidth,trim={0 1.4cm 0 3.4cm},clip]{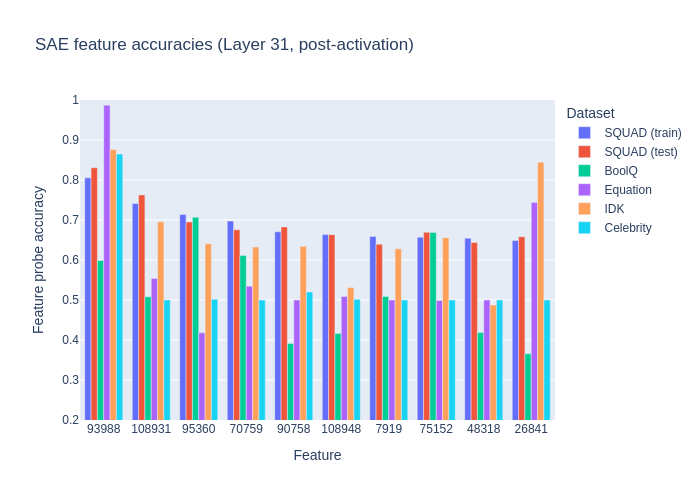}
    \caption{Answerability detection accuracies for top SAE features (Layer 31, post-activation).}
    \label{fig:sae-probe_post31}
\end{figure*}

\begin{figure*}[t]
    \centering
    \includegraphics[width=0.7\textwidth,trim={0 2.2cm 1cm 3.4cm},clip]{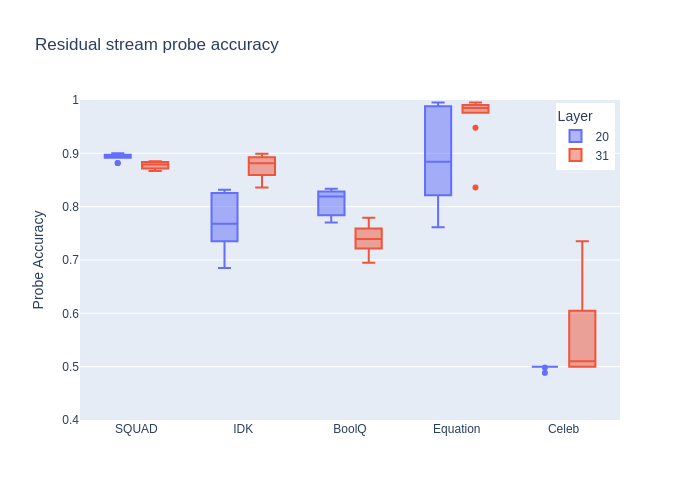}
    \caption{Linear probe trained on Layer 20 and Layer 31 residual stream (SQuAD) and evaluated on IDK, BoolQ, Celebrity, and Equation. The plot shows the median accuracy including the first and third quartile.}
    \label{fig:res-probe-all}
\end{figure*}

We repeat our SAE feature analysis in Layer 20 of the model, as well as providing additional analysis for SAE features activations sampled after the activation function. Figure~\ref{fig:sae-probe_pre20} shows the Layer 20 results using activations sampled before the activation function, while Figures~\ref{fig:sae-probe_post20} and \ref{fig:sae-probe_post31} show analogous results when sampling SAE activations after the activation function. Sampling after the activation reduces the number of relevant features our probe finds, since many features are inactive. However, this does not change the overall results, as we still find features with good generalization performance. 

Figure~\ref{fig:res-probe-all} shows the probing accuracy for the residual stream linear probe for both Layer 20 and 31. The evaluation is repeated across 10 seeds with different training set splits. While the SAE features, as part of the pre-trained autoencoder model, do not heavily depend on the probing dataset, this is not necessarily true for the residual stream probe. The's probe performance across the out-of-distribution datasets varies strongly, indicating that the generalization performance heavily depends on the minor differences in the training data. 

\subsection{Prompt variations} \label{app:prompt-variations}

\begin{figure*}[t]
    \centering
    \includegraphics[width=0.8\textwidth,trim={0 1.4cm 0 3.4cm},clip]{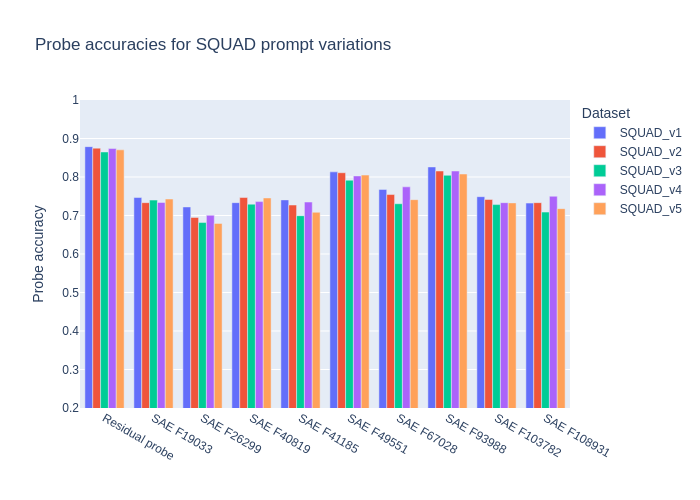}
    \caption{Performance of top SAE features and the residual stream linear probe on variations of prompt used with the SQuAD dataset (layer 31, pre-activation).}
    \label{fig:prompt-variation}
\end{figure*}

\begin{table*}[h]
    \centering
    \begin{tabular}{lp{10cm}}
        \toprule
        Default & Given the following passage and question, answer the question:\newline Passage: \{passage\}\newline Question: \{question\} \\
        \midrule
        Variation 1 & Please read this passage and respond to the query that follows:\newline Passage: \{passage\}\newline Question: \{question\} \\
        \midrule
        Variation 2 & Based on the text below, please address the following question:\newline Text: \{passage\}\newline Question: \{question\} \\
        \midrule
        Variation 3 & Consider the following excerpt and respond to the inquiry:\newline Excerpt: \{passage\}\newline Inquiry: \{question\} \\
        \midrule
        Variation 4 & Review this content and answer the question below:\newline Content: \{passage\}\newline Question: \{question\} \\
        \midrule
        Variation 5 & Using the information provided, respond to the following:\newline Information: \{passage\}\newline Query: \{question\} \\
        \bottomrule
    \end{tabular}
    \caption{SQuAD prompt template variations.}
    \label{table:variations}
\end{table*}


We investigated if the SAE features or the residual stream probes are sensitive to small variations in the prompt. To evaluate this question, we created five variations of the prompt template used for the SQuAD training data (see Table~\ref{table:variations}). The results can be found in Figure~\ref{fig:prompt-variation}, and indicate neither the residual stream probe nor the SAE features are sensitive to this kind of variation. 

\subsection{In-domain SAE feature accuracies} \label{app:in-domain-eval}

\begin{figure*}[t]
    \centering
    \includegraphics[width=0.8\textwidth,trim={0 1.4cm 0 3.4cm},clip]{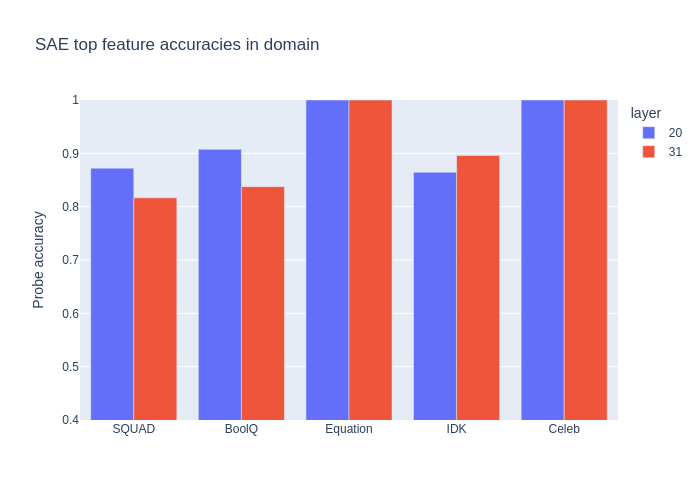}
    \caption{Performance of the top SAE feature's probing accuracy when training and evaluating features on each dataset individually (pre-activation).}
    \label{fig:top-in-domain}
\end{figure*}

Figure~\ref{fig:top-in-domain} shows the accuracy of 1-sparse SAE feature probes for each dataset individually, demonstrating that each of our contrastive datasets is detectable with a probing accuracy of over 80\%.

\subsection{SAE Feature Combination Analyses} \label{app:feature-combinations}

\begin{figure*}[t]
    \centering
    \includegraphics[width=0.8\textwidth,trim={0 1.4cm 0 3.4cm},clip]{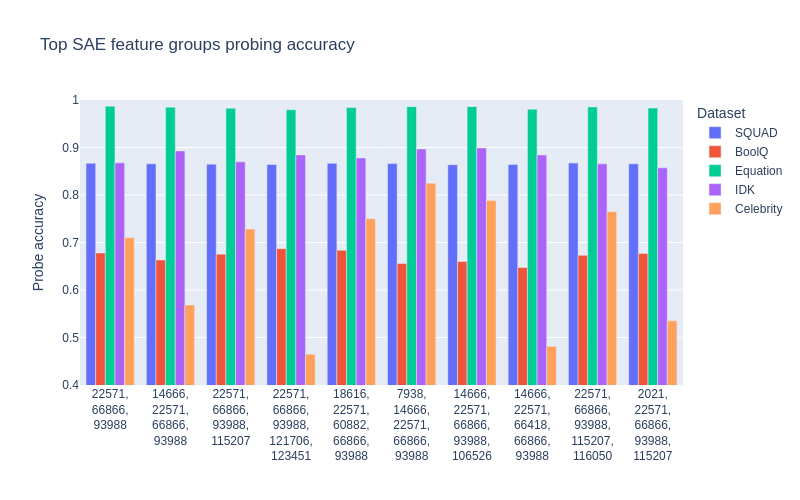}
    \caption{Performance of top feature combinations (layer 31, pre-activation).}
    \label{fig:top-combis}
\end{figure*}

Figure~\ref{fig:top-combis} shows additional probing analysis for the best performing groups of SAE features up to a group size of five. Group performance is generally dominated by the best performing features and does not majorly exceed the performance of the strongest feature. 


\begin{figure*}[t]
    \centering
    \includegraphics[width=0.8\textwidth,trim={0 1.4cm 0 3.4cm},clip]{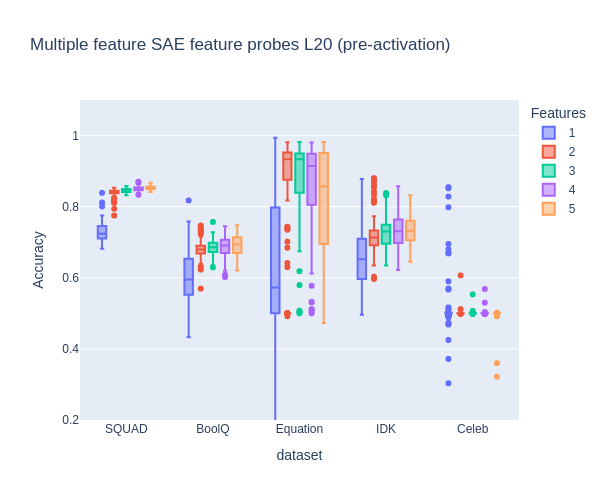}
    \caption{Accuracies of SAE probes trained on different numbers of SAE features (Layer 20, pre-activation).}
    \label{fig:sae-k-pre20}
\end{figure*}

Figure~\ref{fig:sae-k-pre20} shows additional analysis for SAE feature combinations in Layer~20, analogous to the results for Layer~31 given in Figure~\ref{fig:sae-k-pre31}.

\subsection{Cosine Similarities} \label{app:cosine-similarities}
\begin{figure*}[t]
    \centering
    \includegraphics[width=0.8\textwidth,trim={0 0 0 3.4cm},clip]{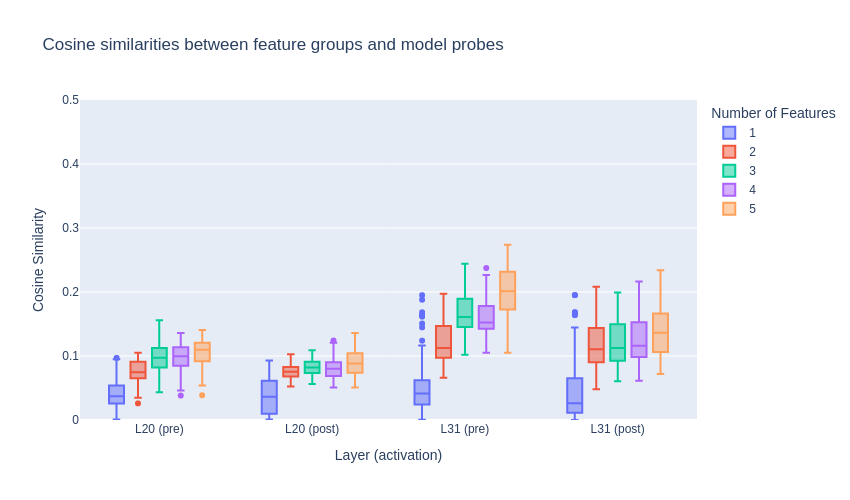}
    \caption{Absolute cosine similarities of top 10 SAE features at different layers, compared with the residual stream probe.}
    \label{fig:similarity_k}
\end{figure*}

We conducted an additional similarity analysis for the top SAE feature groups of different sizes. The results can be found in Figure~\ref{fig:similarity_k} and show a clear trend of larger groups of features becoming more similar to the linear probes. This provides some weak evidence that by default, linear probes might learn more specialized directions that can be represented as a linear combination of more general SAE features. 

\subsection{Spanish language generalization} \label{sec:spanish_eval}

\begin{figure*}[t]
    \centering
    \includegraphics[width=0.8\textwidth,trim={0 0 0 3.4cm},clip]{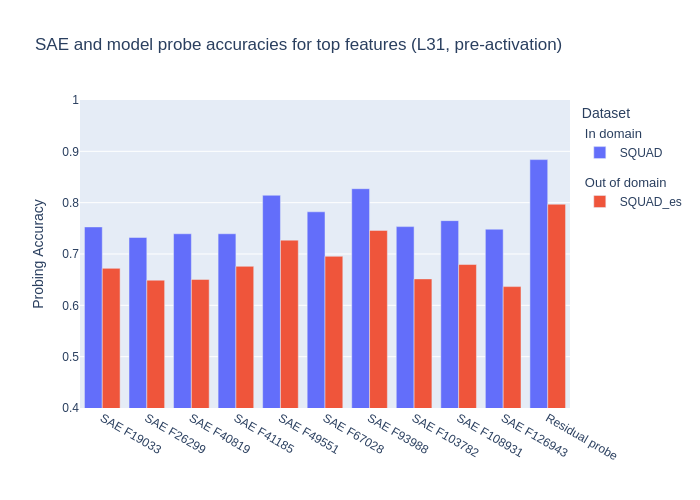}
    \caption{Probing accuracies for top SAE features and the residual stream probe when evaluated on Spanish OOD data.}
    \label{fig:ood_spanish}
\end{figure*}

We performed an additional generalization experiment using a Spanish language variant of SQUAD \cite{carrino2019automatic}. Similar to our other OOD experiments, we used residual and SAE probes trained on the English SQUAD dataset and evaluated their generalization capabilities to the Spanish SQUAD data. Our results did not show large differences in generalization between individual features or the residual stream probe (see Figure~\ref{fig:ood_spanish}). Instead, all top probes and the residual probe generalized roughly equally well, with a performance gap of slightly below 10\%.

\paragraph{Other experiments}
We validated our setup by searching for bias-related features as it was done in related works.
We also experimented with (inofficial) SAEs for an instruction-tuned Llama model, but could not find SAE features with sufficient in-domain probing accuracy. Finally, we also performed analysis on Gemma 2 2B and also the base models, but performance on the answerability task was relatively low in these models (the best SAE features achieved around 70\% probing accuracy).

%% file: latex/02-preliminaries.tex
\section{Preliminaries}\label{app:prelims}


\paragraph{1- sparse SAE probes}
To evaluate how well SAE features predict a certain abstract feature, we utilize 1-sparse probes \cite{gurnee2023finding}. Specifically, we collect activations of a specific SAE feature on a contrastive dataset containing both answerable and not answerable examples, and fit a slope coefficient and intercept to predict the dataset label using linear regression. The Gemma 2 SAEs are trained using a JumpReLU activation function \cite{lieberum2024gemma}. We can sample SAE activations after the activation function (post-relu) or before (pre-relu).
Since there are more learnt features to be found in the latter setting, the main paper figures focus on that. However, we report all results for the post-relu setting in the appendix.

\paragraph{Residual stream probes}

Our residual stream probes are trained on model activations sampled from the model's residual stream. To avoid overfitting, we train the regression model using 5-fold cross validation and perform a hyperparameter optimization by sweeping over regularization parameters with 26 logarithmically spaced steps between 0.0001 and 1. To measure the variability of residual stream probes, we repeat our analysis 10 times with different randomly sampled training datasets.

\paragraph{N-sparse SAE probes}
To train SAE probes with more than 1 feature, we follow the general methodology of our 1-sparse probes. As testing all possible SAE feature combinations is computationally infeasible, we iteratively increase the number of features while testing only the most promising candidates for higher features combinations. Specifically, to find combinations of $k$ features, we use the top 50 best performing features of size $k-1$ and test all possible new combinations with the 500 best performing single SAE features. We use a constant regularization parameter of 1 for the probes, regardless of the number of features.

\paragraph{Feature similarities}

To calculate feature similarities, we use the cosine similarity of the corresponding SAE encoder weight and the slope coefficients of the linear probes trained on the residual stream. SAE features are only compared to other SAE features of the same SAE, and residual stream probes trained at the same location in the model as the SAE. To compare how similar differently sized groups of SAE features are to the residual stream probes, we calculate the mean absolute cosine sim of the top 10 best performing SAE features of a certain group size (1 to 5) with the 10 residual stream probes trained on different training subsets.

%% file: main.bbl
\begin{thebibliography}{33}
\providecommand{\natexlab}[1]{#1}
\providecommand{\url}[1]{\texttt{#1}}
\expandafter\ifx\csname urlstyle\endcsname\relax
  \providecommand{\doi}[1]{doi: #1}\else
  \providecommand{\doi}{doi: \begingroup \urlstyle{rm}\Url}\fi

\bibitem[Barez et~al.(2025)Barez, Fu, Prabhu, Casper, Sanyal, Bibi, O'Gara, Kirk, Bucknall, Fist, Ong, Torr, Lam, Trager, Krueger, Mindermann, Hernandez-Orallo, Geva, and Gal]{Barez2025-open}
Barez, F., Fu, T., Prabhu, A., Casper, S., Sanyal, A., Bibi, A., O'Gara, A., Kirk, R., Bucknall, B., Fist, T., Ong, L., Torr, P., Lam, K.-Y., Trager, R., Krueger, D., Mindermann, S., Hernandez-Orallo, J., Geva, M., and Gal, Y.
\newblock Open problems in machine unlearning for ai safety.
\newblock 2025.
\newblock URL \url{https://arxiv.org/abs/2501.04952}.

\bibitem[Bloom et~al.(2024)Bloom, Tigges, and Chanin]{bloom2024saetrainingcodebase}
Bloom, J., Tigges, C., and Chanin, D.
\newblock Saelens.
\newblock \url{https://github.com/jbloomAus/SAELens}, 2024.

\bibitem[Bricken et~al.(2023)Bricken, Templeton, Batson, Chen, Jermyn, Conerly, Turner, Anil, Denison, Askell, Lasenby, Wu, Kravec, Schiefer, Maxwell, Joseph, Hatfield-Dodds, Tamkin, Nguyen, McLean, Burke, Hume, Carter, Henighan, and Olah]{bricken2023monosemanticity}
Bricken, T., Templeton, A., Batson, J., Chen, B., Jermyn, A., Conerly, T., Turner, N., Anil, C., Denison, C., Askell, A., Lasenby, R., Wu, Y., Kravec, S., Schiefer, N., Maxwell, T., Joseph, N., Hatfield-Dodds, Z., Tamkin, A., Nguyen, K., McLean, B., Burke, J.~E., Hume, T., Carter, S., Henighan, T., and Olah, C.
\newblock Towards monosemanticity: Decomposing language models with dictionary learning.
\newblock \emph{Transformer Circuits Thread}, 2023.
\newblock https://transformer-circuits.pub/2023/monosemantic-features/index.html.

\bibitem[Bricken et~al.(2024)Bricken, Marcus, Mishra-Sharma, Tong, Perez, Sharma, Rivoire, and Henighan]{bricken2024using}
Bricken, T., Marcus, J., Mishra-Sharma, S., Tong, M., Perez, E., Sharma, M., Rivoire, K., and Henighan, T.
\newblock Using dictionary learning features as classifiers.
\newblock \emph{Transformer Circuits Thread}, 2024.
\newblock https://transformer-circuits.pub/2024/features-as-classifiers/index.html.

\bibitem[Carrino et~al.(2019)Carrino, Costa-Juss{\`a}, and Fonollosa]{carrino2019automatic}
Carrino, C.~P., Costa-Juss{\`a}, M.~R., and Fonollosa, J.~A.
\newblock Automatic spanish translation of the squad dataset for multilingual question answering.
\newblock \emph{arXiv preprint arXiv:1912.05200}, 2019.

\bibitem[Chanin et~al.(2024)Chanin, Wilken-Smith, Dulka, Bhatnagar, and Bloom]{chanin2024absorption}
Chanin, D., Wilken-Smith, J., Dulka, T., Bhatnagar, H., and Bloom, J.
\newblock A is for absorption: Studying feature splitting and absorption in sparse autoencoders.
\newblock \emph{arXiv preprint arXiv:2409.14507}, 2024.

\bibitem[Chuang et~al.(2024)Chuang, Qiu, Hsieh, Krishna, Kim, and Glass]{chuang2024lookback}
Chuang, Y.-S., Qiu, L., Hsieh, C.-Y., Krishna, R., Kim, Y., and Glass, J.
\newblock Lookback lens: Detecting and mitigating contextual hallucinations in large language models using only attention maps.
\newblock \emph{arXiv preprint arXiv:2407.07071}, 2024.

\bibitem[Cunningham et~al.(2023)Cunningham, Ewart, Riggs, Huben, and Sharkey]{cunningham2023sparse}
Cunningham, H., Ewart, A., Riggs, L., Huben, R., and Sharkey, L.
\newblock Sparse autoencoders find highly interpretable features in language models.
\newblock \emph{arXiv preprint arXiv:2309.08600}, 2023.

\bibitem[Demircan et~al.(2024)Demircan, Saanum, Jagadish, Binz, and Schulz]{demircan2024sparse}
Demircan, C., Saanum, T., Jagadish, A.~K., Binz, M., and Schulz, E.
\newblock Sparse autoencoders reveal temporal difference learning in large language models.
\newblock \emph{arXiv preprint arXiv:2410.01280}, 2024.

\bibitem[Dubey et~al.(2024)Dubey, Jauhri, Pandey, Kadian, Al-Dahle, Letman, Mathur, Schelten, Yang, Fan, et~al.]{dubey2024llama}
Dubey, A., Jauhri, A., Pandey, A., Kadian, A., Al-Dahle, A., Letman, A., Mathur, A., Schelten, A., Yang, A., Fan, A., et~al.
\newblock The llama 3 herd of models.
\newblock \emph{arXiv preprint arXiv:2407.21783}, 2024.

\bibitem[Elhage et~al.(2021)Elhage, Nanda, Olsson, Henighan, Joseph, Mann, Askell, Bai, Chen, Conerly, DasSarma, Drain, Ganguli, Hatfield-Dodds, Hernandez, Jones, Kernion, Lovitt, Ndousse, Amodei, Brown, Clark, Kaplan, McCandlish, and Olah]{elhage2021mathematical}
Elhage, N., Nanda, N., Olsson, C., Henighan, T., Joseph, N., Mann, B., Askell, A., Bai, Y., Chen, A., Conerly, T., DasSarma, N., Drain, D., Ganguli, D., Hatfield-Dodds, Z., Hernandez, D., Jones, A., Kernion, J., Lovitt, L., Ndousse, K., Amodei, D., Brown, T., Clark, J., Kaplan, J., McCandlish, S., and Olah, C.
\newblock A mathematical framework for transformer circuits.
\newblock \emph{Transformer Circuits Thread}, 2021.
\newblock https://transformer-circuits.pub/2021/framework/index.html.

\bibitem[Gao et~al.(2024)Gao, la~Tour, Tillman, Goh, Troll, Radford, Sutskever, Leike, and Wu]{gao2024scaling}
Gao, L., la~Tour, T.~D., Tillman, H., Goh, G., Troll, R., Radford, A., Sutskever, I., Leike, J., and Wu, J.
\newblock Scaling and evaluating sparse autoencoders.
\newblock \emph{arXiv preprint arXiv:2406.04093}, 2024.

\bibitem[Gurnee et~al.(2023)Gurnee, Nanda, Pauly, Harvey, Troitskii, and Bertsimas]{gurnee2023finding}
Gurnee, W., Nanda, N., Pauly, M., Harvey, K., Troitskii, D., and Bertsimas, D.
\newblock Finding neurons in a haystack: Case studies with sparse probing.
\newblock \emph{arXiv preprint arXiv:2305.01610}, 2023.

\bibitem[Kantamneni et~al.(2024)Kantamneni, Engels, Rajamanoharan, and Nanda]{kantamneni2024saeprobing}
Kantamneni, S., Engels, J., Rajamanoharan, S., and Nanda, N.
\newblock Sae probing: What is it good for? absolutely something!
\newblock \url{https://www.lesswrong.com/posts/NMLq8yoTecAF44KX9/sae-probing-what-is-it-good-for-absolutely-something}, 2024.

\bibitem[Kantamneni et~al.(2025)Kantamneni, Engels, Rajamanoharan, Tegmark, and Nanda]{kantamneni2025sparseautoencodersusefulcase}
Kantamneni, S., Engels, J., Rajamanoharan, S., Tegmark, M., and Nanda, N.
\newblock Are sparse autoencoders useful? a case study in sparse probing, 2025.
\newblock URL \url{https://arxiv.org/abs/2502.16681}.

\bibitem[Kissane et~al.(2024)Kissane, Krzyzanowski, Bloom, Conmy, and Nanda]{kissane2024interpreting}
Kissane, C., Krzyzanowski, R., Bloom, J.~I., Conmy, A., and Nanda, N.
\newblock Interpreting attention layer outputs with sparse autoencoders.
\newblock \emph{arXiv preprint arXiv:2406.17759}, 2024.

\bibitem[Lan et~al.(2024)Lan, Torr, Meek, Khakzar, Krueger, and Barez]{lan2024sparse}
Lan, M., Torr, P., Meek, A., Khakzar, A., Krueger, D., and Barez, F.
\newblock Sparse autoencoders reveal universal feature spaces across large language models, 2024.
\newblock Preprint, arXiv:2410.06981.

\bibitem[Lieberum et~al.(2024)Lieberum, Rajamanoharan, Conmy, Smith, Sonnerat, Varma, Kram{\'a}r, Dragan, Shah, and Nanda]{lieberum2024gemma}
Lieberum, T., Rajamanoharan, S., Conmy, A., Smith, L., Sonnerat, N., Varma, V., Kram{\'a}r, J., Dragan, A., Shah, R., and Nanda, N.
\newblock Gemma scope: Open sparse autoencoders everywhere all at once on gemma 2.
\newblock \emph{arXiv preprint arXiv:2408.05147}, 2024.

\bibitem[Makelov et~al.(2024)Makelov, Lange, and Nanda]{makelov2024towards}
Makelov, A., Lange, G., and Nanda, N.
\newblock Towards principled evaluations of sparse autoencoders for interpretability and control.
\newblock \emph{arXiv preprint arXiv:2405.08366}, 2024.

\bibitem[Marks et~al.(2024{\natexlab{a}})Marks, Paren, Krueger, and Barez]{marks2024enhancing}
Marks, L., Paren, A., Krueger, D., and Barez, F.
\newblock Enhancing neural network interpretability with feature-aligned sparse autoencoders, 2024{\natexlab{a}}.
\newblock Preprint, arXiv:2411.01220.

\bibitem[Marks et~al.(2024{\natexlab{b}})Marks, Rager, Michaud, Belinkov, Bau, and Mueller]{marks2024sparse}
Marks, S., Rager, C., Michaud, E.~J., Belinkov, Y., Bau, D., and Mueller, A.
\newblock Sparse feature circuits: Discovering and editing interpretable causal graphs in language models.
\newblock \emph{arXiv preprint arXiv:2403.19647}, 2024{\natexlab{b}}.

\bibitem[Nanda \& Bloom(2022)Nanda and Bloom]{nanda2022transformerlens}
Nanda, N. and Bloom, J.
\newblock Transformerlens.
\newblock \url{https://github.com/TransformerLensOrg/TransformerLens}, 2022.

\bibitem[Olah et~al.(2018)Olah, Mordvintsev, and Schubert]{olah2018building}
Olah, C., Mordvintsev, A., and Schubert, L.
\newblock The building blocks of interpretability.
\newblock \emph{Distill}, 2018.
\newblock URL \url{https://distill.pub/2018/building-blocks/}.

\bibitem[Olsson et~al.(2022)Olsson, Elhage, Nanda, Joseph, DasSarma, Henighan, Mann, Askell, Bai, Chen, et~al.]{olsson2022context}
Olsson, C., Elhage, N., Nanda, N., Joseph, N., DasSarma, N., Henighan, T., Mann, B., Askell, A., Bai, Y., Chen, A., et~al.
\newblock In-context learning and induction heads.
\newblock \emph{arXiv preprint arXiv:2209.11895}, 2022.

\bibitem[Rajamanoharan et~al.(2024)Rajamanoharan, Conmy, Smith, Lieberum, Varma, Kram{\'a}r, Shah, and Nanda]{rajamanoharan2024improving}
Rajamanoharan, S., Conmy, A., Smith, L., Lieberum, T., Varma, V., Kram{\'a}r, J., Shah, R., and Nanda, N.
\newblock Improving dictionary learning with gated sparse autoencoders.
\newblock \emph{arXiv preprint arXiv:2404.16014}, 2024.

\bibitem[Rajpurkar et~al.(2018)Rajpurkar, Jia, and Liang]{rajpurkar2018know}
Rajpurkar, P., Jia, R., and Liang, P.
\newblock Know what you don't know: Unanswerable questions for squad.
\newblock \emph{arXiv preprint arXiv:1806.03822}, 2018.

\bibitem[Sclar et~al.(2023)Sclar, Choi, Tsvetkov, and Suhr]{sclar2023quantifying}
Sclar, M., Choi, Y., Tsvetkov, Y., and Suhr, A.
\newblock Quantifying language models' sensitivity to spurious features in prompt design or: How i learned to start worrying about prompt formatting.
\newblock \emph{arXiv preprint arXiv:2310.11324}, 2023.

\bibitem[Slobodkin et~al.(2023)Slobodkin, Goldman, Caciularu, Dagan, and Ravfogel]{slobodkin2023curious}
Slobodkin, A., Goldman, O., Caciularu, A., Dagan, I., and Ravfogel, S.
\newblock The curious case of hallucinatory (un) answerability: Finding truths in the hidden states of over-confident large language models.
\newblock In \emph{Proceedings of the 2023 Conference on Empirical Methods in Natural Language Processing}, pp.\  3607--3625, 2023.

\bibitem[Sulem et~al.(2021)Sulem, Hay, and Roth]{sulem2021we}
Sulem, E., Hay, J., and Roth, D.
\newblock Do we know what we don’t know? studying unanswerable questions beyond squad 2.0.
\newblock In \emph{Findings of the Association for Computational Linguistics: EMNLP 2021}, pp.\  4543--4548, 2021.

\bibitem[Sulem et~al.(2022)Sulem, Hay, and Roth]{sulem2022yes}
Sulem, E., Hay, J., and Roth, D.
\newblock Yes, no or idk: The challenge of unanswerable yes/no questions.
\newblock In \emph{Proceedings of the 2022 Conference of the North American Chapter of the Association for Computational Linguistics: Human Language Technologies}, pp.\  1075--1085, 2022.

\bibitem[Team et~al.(2024)Team, Riviere, Pathak, Sessa, Hardin, Bhupatiraju, Hussenot, Mesnard, Shahriari, Ram{\'e}, et~al.]{team2024gemma}
Team, G., Riviere, M., Pathak, S., Sessa, P.~G., Hardin, C., Bhupatiraju, S., Hussenot, L., Mesnard, T., Shahriari, B., Ram{\'e}, A., et~al.
\newblock Gemma 2: Improving open language models at a practical size.
\newblock \emph{arXiv preprint arXiv:2408.00118}, 2024.

\bibitem[Templeton et~al.(2024)Templeton, Conerly, Marcus, Lindsey, Bricken, Chen, Pearce, Citro, Ameisen, Jones, Cunningham, Turner, McDougall, MacDiarmid, Freeman, Sumers, Rees, Batson, Jermyn, Carter, Olah, and Henighan]{templeton2024scaling}
Templeton, A., Conerly, T., Marcus, J., Lindsey, J., Bricken, T., Chen, B., Pearce, A., Citro, C., Ameisen, E., Jones, A., Cunningham, H., Turner, N.~L., McDougall, C., MacDiarmid, M., Freeman, C.~D., Sumers, T.~R., Rees, E., Batson, J., Jermyn, A., Carter, S., Olah, C., and Henighan, T.
\newblock Scaling monosemanticity: Extracting interpretable features from claude 3 sonnet.
\newblock \emph{Transformer Circuits Thread}, 2024.
\newblock URL \url{https://transformer-circuits.pub/2024/scaling-monosemanticity/index.html}.

\bibitem[Yun et~al.(2021)Yun, Chen, Olshausen, and LeCun]{yun2021transformer}
Yun, Z., Chen, Y., Olshausen, B.~A., and LeCun, Y.
\newblock Transformer visualization via dictionary learning: contextualized embedding as a linear superposition of transformer factors.
\newblock \emph{arXiv preprint arXiv:2103.15949}, 2021.

\end{thebibliography}
